\documentclass[letterpaper, 10 pt, conference]{ieeeconf}

\IEEEoverridecommandlockouts                              
\usepackage{amsmath,amsfonts}
\usepackage{array}
\usepackage[caption=false,font=normalsize,labelfont=sf,textfont=sf]{subfig}
\usepackage{textcomp}
\usepackage{stfloats}
\usepackage{url}
\usepackage{verbatim}

\usepackage{bm}
\usepackage{inputenc}
\usepackage{graphicx}
\usepackage{amssymb}
\usepackage{epstopdf}
\usepackage{multicol}
\usepackage{cite}
\usepackage{xcolor}
\usepackage{algorithmicx}
\usepackage[ruled,norelsize]{algorithm2e}

\newenvironment{changemargin}[2]{%
\begin{list}{}{%
\setlength{\topsep}{0pt}%
\setlength{\leftmargin}{#1}%
\setlength{\rightmargin}{#2}%
\setlength{\listparindent}{\parindent}%
\setlength{\itemindent}{\parindent}%
\setlength{\parsep}{\parskip}%
}%
\item[]}{\end{list}}

\makeatletter
\newcommand{\removelatexerror}{\let\@latex@error\@gobble}
\makeatother

\graphicspath{{./images/}}

\usepackage{color}

\title{\LARGE \bf
Adaptive Gait Modeling and Optimization for \\ Principally Kinematic Systems
}

\author{Siming Deng$^{1,2}$, Noah J.~Cowan$^{1,2,\dagger}$, Brian A.~Bittner$^{3,\dagger}$
\thanks{This material is based upon work supported by the National Science Foundation under grants no.\ \#1830893, EFRI C3 SoRo: Programming Thermobiochemomechanical (TBCM) Multiplex Robot Gels}%
\thanks{$^{1}$Laboratory for Computational Sensing and Robotics, Johns Hopkins University,  Baltimore MD 21218 USA.}
\thanks{$^{2}$Department of Mechanical Engineering, Johns Hopkins University, Baltimore MD 21218 USA.}
\thanks{$^{3}$Johns Hopkins University Applied Physics Lab, Laurel,
  MD, 97331  USA.}
\thanks{$^{\dagger}$Co-supervised equally.}
}

\begin{document}

\maketitle
\thispagestyle{empty}
\pagestyle{empty}

%%%%%%%%%%%%%%%%%%%%%%%%%%%%%%%%%%%%%%%%%%%%%%%%%%%%%%%%%%%%%%%%%%%%%%%%%%%%%%%%
\begin{abstract}
  Robotic adaptation to unanticipated operating conditions is crucial to achieving persistence and robustness in complex real world settings.  For a wide range of cutting-edge robotic systems, such as micro- and nano-scale robots, soft robots, medical robots, and bio-hybrid robots, it is infeasible to anticipate the operating environment \emph{a priori} due to complexities that arise from numerous factors including imprecision in manufacturing, chemo-mechanical forces, and poorly understood contact mechanics.  Drawing inspiration from data-driven modeling, geometric mechanics (or gauge theory), and adaptive control, we employ an adaptive system identification framework and demonstrate its efficacy in enhancing the performance of principally kinematic locomotors (those governed by Rayleigh dissipation or zero momentum conservation). We showcase the capability of the adaptive model to efficiently accommodate varying terrains and iteratively modified behaviors within a behavior optimization framework.  This provides both the ability to improve fundamental behaviors and perform motion tracking to precision.  Notably, we are capable of optimizing the gaits of the Purcell swimmer using approximately 10 cycles per link, which for the nine-link Purcell swimmer provides a factor of ten improvement in optimization speed over the state of the art. Beyond simply a computational speed up, this ten-fold improvement may enable this method to be successfully deployed for \textit{in-situ} behavior refinement, injury recovery, and terrain adaptation, particularly in domains where simulations provide poor guides for the real world.
\end{abstract}
%%%%%%%%%%%%%%%%%%%%%%%%%%%%%%%%%%%%%%%%%%%%%%%%%%%%%%%%%%%%%%%%%%%%%%%%%%%%%%%%

\section{Introduction}

% \begin{outline}[gray]
%   \item General robotics, motivation for adaptive modeling and control
%   \item Related work: data-driven methods, (adaptive system id), gait design 
%   \item Contributions: adaptive filters in place of batch regression, applications: in-situ gait design, adaptation to changing environments (e.g., terrain, payload, etc.)
% \end{outline}

The field of robotics has experienced significant advancements over the past few decades, evolving from systems designed for explicit situations into more generalizable and versatile machines.
This transformation has been driven by the ever-growing demand for robots that are capable of interacting with dynamic and unstructured environments.
In many such environments, learning from \textit{in-situ} experiences is crucial to functioning in the field.
These challenges are particularly evident for systems with high internal degrees of freedom and complex system-environment interactions, where subtle changes in environmental characteristics can drive fundamental differences in emergent behavior.

Current paradigms of motion control are not ideally suited to attaining precision in these unstructured settings.  Traditional \textit{bottom-up}, model predictive approaches to robotic control heavily rely on explicit accounting of physical quantities which can be derived from first principles or hand-crafted to the desired precision of the user \cite{grandia2021multi,gong2022zero}.  Adaptive system identification extends models to react to anticipated changes in the model.  The rate and quality of adaptation are affected strongly by the assumptions about observable perturbations of the model \cite{weingarten2004automated, brambilla2006adaptive, sun2016adaptive}.  In general, these methods are ineffective when either (a) the range of assumed adaptations is so large that overfitting occurs or (b) the assumptions are too specific such that violations of the assumptions are common.  More recent approaches to motion control include reinforcement learning and deep learning approaches, both of which require large, infrastructurally cumbersome quantities of training data (simulated or real) \cite{peng2017deeploco,heess2017emergence,tan2018sim}.  Here, the robot is particularly ill-suited to succeed in unanticipated scenarios, since (by definition) they would not be included in the training set.  Efforts in domain randomization have yielded reinforcement learning agents more capable of generalization and sim-to-real transfer, but often at the expense of achieving precise mastery of individual domains \cite{tobin2017domain,mullins2018adaptive}. Learning frameworks such dynamic mixtures of experts \cite{rambabu2019mixture,tsuda2020modeling} attempt to achieve robustness and precision with distributed expertise, but the computation for these experts to both decouple from each others' domains and refine individual expertise has not yet been designed for real-time adaptation in complex fielded environments. Adaptive control, reinforcement learning, and deep learning architectures are not designed to rapidly and precisely handle unanticipated changes to the dynamics.

This limits our ability to advance robots to new application spaces that can handle complexities such as customized body parts, imprecise fabrications, and poorly characterized environments.  Such considerations are crucial for soft robots \cite{rozen2021design,pantula2022untethered}, nano-robots \cite{li2020progress}, medical robots \cite{park2022computational}, and bio-hybrid robots \cite{webster2022biohybrid}, as each are subject to these considerations.  In this work, we use an adaptive system identification framework to leverage the \textit{top-down} modeling insights of geometric mechanics \cite{Bittner2018}.  This permits a framework for adaptation that captures the general space of possible dynamics changes within a class of physics, but it is representationally compact enough to be fit rapidly to data \textit{in situ}.  The result enables adapting motion models to distinctly new regimes without knowing \emph{a priori} the nature of that change (e.g. a substrate change, morphology change, or weight distribution change).  The assumptions made are that the dissipative nature of the physics persists and that the recent history of the behavior is evolving in a homogeneous environment. The key contribution of our work is that empirical modeling is enhanced to update dynamically in subsecond time with new observations (shape and body velocity), permitting rapid adaptation to abrupt, unanticipated changes in robot morphology, weight distribution, or substrate conditions.

We restrict our attention in this work to systems that are principally kinematic \cite{kelly1995geometric}, a property developed in the geometric mechanics community that describes the locomotion of systems governed by dissipative forces or conservation of zero momentum.  These systems enjoy an exploitable reduction from second-order mechanical dynamics to ones that possess a first-order structure, namely a linear mapping from internal configuration (shape) velocity to body velocity.  Recent work has shown that this reduction provides a reduced computational structure for fitting robot models that are broadly descriptive of many classes of animal and robot behaviors \cite{zhao2022walking,chong2023self}.  These works have established that many robot and animal behaviors are describable by the space of dissipative models, which span a dense and broad variety of animal morphologies, weight distributions, and environmental substrates.  Bittner et al. \cite{Bittner2018} introduced a data-driven method for constructing a local model in the neighborhood of an observed gait.  This method relies on data points collected from stochastically perturbed and iteratively repeated behaviors.  The ability to construct local models makes it possible to sample candidate gaits offline, which provides an opportunity for sample efficient gait optimization as well as gait modulations for continuous gait control \cite{dengenhancing2022}.

% Adaptive modeling refers to the capacity of robotic systems to continuously update their internal representations, behaviors, and strategies in response to changing circumstances, unforeseen challenges, and novel tasks.
% Adaptive modeling harnesses the power of machine learning, artificial intelligence, and data-driven techniques to endow robots with the capability to adapt and improve their performance autonomously.
% It enables robots to acquire knowledge from their interactions with the world, make informed decisions based on past experiences, and refine their skills over time.
% This transition from rule-based programming to adaptive modeling has paved the way for the development of robots that can thrive in real-world scenarios where uncertainty, variability, and change are the norm rather than the exception.

% \njccom{I'm removing ``extend / extension where I can as it sounds 'incremental'.}

Here, we capitalize on the use of adaptive filters in constructing the aforementioned local models in real time. The advantages of this reframing are multifold. In this continuously updating paradigm, the model automatically adapts to changes in physical interactions dynamically, rather than awaiting a batch update as in prior work. This actively updated model enables tracking of the model confidence, which is crucial in determining how much empirical data is required to obtain a model with acceptable prediction quality for behavior refinement and control. We show that the rate and quality of adaptation permit refinements of behaviors to master unanticipated environments with speed and precision. We discuss that as internal complexity is added to the system, we encounter situations where engineering decisions between speed and reliable precision need to be made.

Next, we will provide technical background on the adaptive filters, followed by a description of the filter which takes a recursive formulation. We follow this with several experiments that demonstrate the performance of the adaptive filter, noting its accuracy in capturing the dynamics of the principally kinematic Purcell swimmer through a regime change from low drag ratio to high drag ratio swimming. Finally, we offer points of discussion and our conclusions on the performance and implications of the proposed approach.

\section{Background}

The mathematical framework used to capture the motion model of principally kinematic systems is covered here. We also discuss prior efforts to model these systems from data and their current limitations. 

\subsection{Principally Kinematic Systems}

The field of geometric mechanics \cite{cendra2001geometric} \cite{BlochNonholonomic2003} \cite{ostrowskioptimal1998} has pursued the applications of symmetries to mechanics.
In many cases, these applications result in reduced computational structures for system representation and analysis.
One such insight is the reduced Lagrangian or reconstruction equation, which writes the mechanics of the systems using three distinct equations describing the \textit{generalized momentum}, body velocity, and shape dynamics of the system.
Notably, these equations describe the interactions of general second-order systems in ways that one can infer in fundamentally different ways than classical rigid body dynamics formulations.
This representation leans on a key symmetry that the environment remains homogeneous, such that, for example, the body velocity equation remains a constant function of shape, shape velocity, and momentum, no matter where you place the object in the environment.

In this work, we will pursue an adaptation scheme that applies to a subclass of these systems. 
Here, we focus on principally kinematic systems. 
Commonly, these can be observed as systems where dissipative forces dominate the dynamics \cite{kelly1996geometry}.
Recent work has suggested this class of dissipation model, despite its simplicity, is highly descriptive of a broad class of robots and animals \cite{zhao2022walking,chong2023self}.
Principally kinematic systems are also observed where conservation of zero momentum drives the physics \cite{marsden1998symmetries}.
These scenarios are less commonly observed but applicable to orbital manipulation.
In principally kinematic systems, the equations of motion can be reduced, resulting in a shape-defined linear mapping from shape velocity to body velocity.
The resulting model is
\begin{equation}
  \label{eq_reconstruction}
  \xi = -\bm{A}(r)\dot{r},
\end{equation}
where $\xi$ is the body velocity, $r$ is the internal configuration, and $\bm{A}(r)$ is the linear mapping termed a \textit{connection}.
In situations concerning momentum conservation, we use the phrasing \textit{mechanical connection}, whereas in dissipative regimes the term \textit{viscous connection} is applied.

\subsection{Data-driven methods for geometric systems} 
\label{datadriven} 
Bittner et al.~\cite{Bittner2018} developed a data-driven approach to geometric modeling and optimization of these principally kinematic systems. 
The approaches provide \textit{in-situ} system identification of various classes of platforms by locally modeling behaviors by observing approximately repeated cycles of the behavior. 
The variance in the repetition provides regularly sampled time series data that can be fit to a generalized oscillator such that all observed samples are assigned an asymptotic phase.
Within phase windows, a linearization of the connection is fit to this data, organized with respect to an average behavior or limit cycle fit using the asymptotic phase.
Across these phase windows, model coefficients provide input to a Fourier series fit which provides a linearized model that can be sampled as a function of phase.
Perturbations from the average behavior are computed as $\delta := r - \theta$, where $\theta$ provides a shape on the limit cycle as a function of phase.
The first-order Taylor approximation of the connection is computed as
\begin{equation}
  \label{eq_bvmodel}
  \bm{A}_k(r) \approx \bm{A}_k(\theta) + \delta^T\frac{\partial \bm{A}_k}{\partial r},
\end{equation}
where $k$ denotes rows of the connection matrix. 

Samples are grouped into overlapping phase bins. 
The size of these bins is a heuristic that can strongly affect model quality. 
Large phase bins give the data set larger insight into dynamics farther along or behind in the cycle. The result of large phase bins is often less overfitting globally, but also less accuracy locally in the phase bin of interest. 
The regression takes the form
\begin{equation}
  \label{eq_glm}
  % \xi_k \sim \bm{A}_k(\theta)\dot{\theta} + \delta^T\frac{\partial \bm{A}_k}{\partial r}\dot{\theta} + \bm{A}_k(\theta)\dot{\delta} + \delta^T\frac{\partial \bm{A}_k}{\partial r}\dot{\delta}.
  \xi^{(n)}_k \sim \bm{C}_k + \bm{B}_k\delta^{(n)} + \bm{A}_k(\theta)\dot{\delta}^{(n)} + \delta^{T(n)}\frac{\partial \bm{A}_k}{\partial r}\dot{\delta}^{(n)},
\end{equation}
where $k$ provides the output dimension, $n$ is the sample number. These regressions take place in phase windows centered about a phase-defined shape on the limit cycle, $\theta_m$.
This model is fitted in phase windows across the cycle and then fitted to a Fourier series, providing a phase-queried linearization of approximately repeated behaviors.

The initial limitations of this approach involved its applicability to systems with momentum. Follow-up work was done to capture and model the behavior of systems with dissipating momentum while maintaining a first-order prediction structure \cite{kvalheim2019gait}. 
Additional follow-on work extended these models to principally kinematic systems that do not possess high bandwidth control of their shape dynamics. 
Initial extensions addressed systems with high bandwidth control of a subset of the entire shape space \cite{Bittnersuds2022} but eventually worked towards addressing systems with low bandwidth control of shape variables \cite{deng2023data}.
In this paper, we explore an extension to the initial work \cite{Bittner2018} that permits real-time updates for \emph{in-situ} modeling and adaptation in the field. 
Simultaneously, we work directly towards the overall architecture that can support these extensions into adjacent geometric domains.

\section{Methods}

% \begin{outline}[gray]
%   \item Adaptive filters: RLS
%   \item Adaptive gait design
% \end{outline}

The two core methods of our implementation are presented here. 
One method involves an adaptive filter for data-driven geometric mechanics that takes a recursive formulation such that it is updatable in real-time for fielded platforms. 
The second method covers prediction quality metrics, also computable in real-time, which provide guides for when these adaptive models are suitable for trajectory tracking or informing behavior refinements.
We also cover a stochastic perturbation scheme for generating approximately repeated behaviors, which is the means by which we generate local behavior data for modeling.

\subsection{Adaptive Filters}

Our method employs adaptive filters, which are widely used in system identification.
These filters are designed to estimate the parameters of a linear model by recursively updating the weights as new data samples are available.
The main usage of adaptive filters in system identification is that they can track time-varying systems and adapt to changes in the underlying process.
Recursive Least Square (RLS) filters are a class of adaptive filters that specifically minimize the mean square error of the predicted filter output and the actual system output (measurements) \cite{hayes1996statistical}. Instead of collecting samples in batches and then fitting a model, we designed a cluster of adaptive filters on top of the data-driven modeling framework so that the model can be updated simultaneously as samples are collected.
RLS filters fit the role directly in the linear regression problem in \eqref{eq_glm}. 

In this work, we restrict our attention to approximately cyclic behaviors.
For such behaviors, phase is a useful tool through which to organize behavioral data into localized partitions.
In phase coordinate, namely $S^1$, the nominal gait is written as $\theta(\cdot)$; and phase windows with the same width are set up equally spaced along $S^1$. 
The phase window indexed at $m$ is defined as $[\phi^m-\frac{\Delta\phi}{2}, \phi^m+\frac{\Delta\phi}{2}]$, centered at $\phi^m$ with a width of $\Delta\phi$. 
The nominal gait at the $m^\mathrm{th}$ window center is denoted $\theta^m$ with its derivative $\dot{\theta}^m$.  
In each phase window, an RLS filter is constructed for each system output, body velocities $\xi_k$.
For example, a system operating on $\mathrm{SE}(2)$ should have three independent RLS filters in the same phase window, corresponding to its three-dimensional body velocity. 
Every time a new sample is collected, all filters whose corresponding phase windows cover the sample's phase value are updated as follows:
\vspace{-1em}
\begin{equation}
  \label{eq_rls}
  % y_k = [1, \delta, \dot{\delta}, \delta \otimes \dot{\delta}] \bm{w}_k,
  \begin{aligned}
    \begin{bmatrix}
      \vdots \\
      \xi_k \\
      \xi_k \\
      \xi_k \\
      \vdots
    \end{bmatrix}
  \end{aligned} = 
  \begin{aligned}
    \begin{bmatrix}
      & & \vdots \\
      1, & \delta^{m-1}, & \dot{\delta}^{m-1}, & \delta^{m-1} \otimes \dot{\delta}^{m-1} \\
      1, & \delta^m \quad, & \dot{\delta}^m \quad, & \delta^m \otimes \dot{\delta}^m \\
      1, & \delta^{m+1}, & \dot{\delta}^{m+1}, & \delta^{m+1} \otimes \dot{\delta}^{m+1} \\
      & & \vdots
    \end{bmatrix}
  \end{aligned} \cdot
  \begin{aligned}
    \begin{bmatrix}
      \vdots \\
      \bm{w}_k^{m-1} \\
      \bm{w}_k^m \\
      \bm{w}_k^{m+1} \\ 
      \vdots
    \end{bmatrix}^T
  \end{aligned}
\end{equation}

Here, $\xi_k$ is the desired value of the $k^\mathrm{th}$ body velocity, $\delta^m:=r-\theta^m, \dot{\delta}^m:=\dot{r}-\dot{\theta}^m$ are the shape and shape velocity perturbation samples defined in the $m^\mathrm{th}$ phase window.
 $\bm{w}_k^m$ is the weight vector of the $k^\mathrm{th}$ RLS filter in the $m^\mathrm{th}$ phase window, where the dimension is $1+2d+d^2$, $d$ being the dimension of the system shape.
The regressor matrix is constructed by stacking the shape and shape velocity perturbation samples in all phase windows that cover the sample's phase value.
% 
% \begin{figure}[!t]
%   \removelatexerror
%   \begin{algorithm}[H]
%   \caption{adaptive filter update}
%   \label{alg:adaptive_filter}
%     \While(update all nearby filters){new sample collected}{
%       list\_filters = get\_filters\_to\_update(sample.phase)
%       \For( ){filter in list\_filters}{
%         filter.update(sample)
%       }
%     }
%   \end{algorithm}
% \end{figure}
% 
The full model of the system is constructed by combining all the filter weights and smoothed along $\phi$ by a Fourier series.

An important parameter in the RLS filter algorithm is the forgetting factor $\lambda_{RLS} \in (0,1)$, which gives exponentially less weight to older samples.
The forgetting factor, in combination with other parameters (e.g. phase window width and spacing, data sampling rate in terms of phase), determines how much data is effectively in the memory of the model.
This number can be tuned for the robustness vs. adaptability trade-off, where larger memory results in a more stable model while being slower to adapt to real changes.

\subsection{Model Metrics}
\label{sec:metrics}

% Prediction error, calculated as the difference between the predicted body velocity and its measured value in the sample, is recorded for each sample. 

We evaluate the first-order data-driven model by comparing its prediction error to that of the baseline phase model as in prior work \cite{Bittnersuds2022}:
\begin{equation}
  \label{eq_metric}
  \Gamma = 1-\frac{\sum^\mathcal{N}_{n=1}\|\xi_D^{(n)}-\xi^{(n)}\|}{\sum^\mathcal{N}_{n=1}\|\xi_T^{(n)}-\xi^{(n)}\|}.
\end{equation}
$\xi^{(n)}$ is the body velocity from the $n^{\mathrm{th}}$ collected sample, $\xi_D^{(n)}$ and $\xi_T^{(n)}$ denote the data-driven and phase (baseline comparison) model predictions for this velocity.
The baseline phase model is effectively the zeroth order approximation, in other words, phase averaged behavior of the seen data. 

In an online situation, it is important to keep track of a measure that gives a level of confidence to the model.
This moving metric can be used as a criterion for determining whether the model has converged and can be used with confidence.
 This metric is actively updated as new samples are collected as the following,
\begin{equation}
  \label{eq_metric_filter}
  \begin{aligned} 
    \psi_{D}^{(n)} &= \lambda_{\Gamma} \cdot \psi_{D}^{(n-1)} + \|\xi_D^{(n)}-\xi^{(n)}\| \\
    \psi_{T}^{(n)} &= \lambda_{\Gamma} \cdot \psi_{T}^{(n-1)} + \|\xi_T^{(n)}-\xi^{(n)}\| \\
    % \Gamma_{\xi}^{(n)} &= 1-\frac{\psi^{(n)}}{\psi^{(n)}}.
    \Gamma &= 1-\frac{\psi_{D}^{(n)}}{\psi_{T}^{(n)}}.
  \end{aligned} 
  % \Gamma_{\xi}^{(n)} = 1-\frac{\lambda_{\Gamma} \cdot \Delta\xi_D^{(n-1)} + \|\xi_D^{(n)}-\xi^{(n)}\|}{\lambda_{\Gamma} \cdot \Delta\xi_T^{(n-1)} + \|\xi_T^{(n)}-\xi^{(n)}\|}.
\end{equation}
Here, $\lambda_{\Gamma} \in (0,1)$ is the forgetting factor, which is set to be nearly 1.
This number determines how heavily new samples are weighted towards the measure and how fast the history of the metric fades away.
Similar to the forgetting factor $\lambda_{RLS}$ in the RLS algorithm, $\lambda_{\Gamma}$ (together with other parameters) effectively controls how much history the metric $\Gamma$ is carrying. For convenience, we will later refer a relatively `rapid' forgetting factor to $\lambda$'s further away from 1, and a `slow' forgetting factor to $\lambda$'s closer to 1.

% \sdcom{TODO: add metric filter details}

\subsection{Online Noise Generation}
\label{sec:noise}

The data-driven modeling framework needs to have good coverage of both shape and shape velocity around the nominal gait.
Hence, we need to generate perturbations in the neighborhood of the nominal gait while commanding the system in real-time. 

To ensure the relative smoothness of the generated shape trajectory, we employed a second-order stochastic differential equation (SDE) as a smoothing filter,
\begin{equation}
  \label{eq_noise}
  % d\delta = -(\alpha \delta)dt + \eta \circ dW
  d\dot{\delta} = -(\alpha \dot{\delta} + \beta \delta)dt + \eta \cdot dW
\end{equation}
where $\eta$ is the brownian noise amplifier coefficient, $\alpha$ and $\beta$ are the attraction coefficients that are kept positive, and $dW$ is the Wiener process.
The noise generated by this SDE is a smoothed first-order perturbation on the system shape velocity, which also ensures that the resulting shape trajectory is pulled back to the nominal gait over time.

% \begin{figure}
%   \centering
%   \includegraphics[width=\columnwidth]{rls_batch_comp.png}
%   \caption{body velocity prediction error comparison between adaptive model and batch model over phase}
%   \label{fig:2dof_rls_batch}
% \end{figure}

\begin{figure}[tb]
  \begin{center}
    \vspace{2mm}
    \includegraphics[width=0.9\columnwidth]{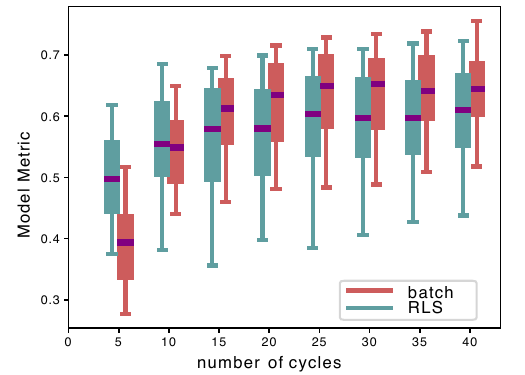}
  \end{center}
  \caption{The Purcell swimmer incrementally experiences more stochastically perturbed cycles in a new operating environment. Model metrics of the batch model (green) and adaptive model (red) are evaluated for their predictive performance at that point in the experiment on a hold-out test set of 40 cycles (for 100 pairs of training and testing trials). The model metrics for both methods are shown in boxplots where each box corresponds to $5,25,50,75,95$ percentiles of a model trained with the first $n \in [5,10,15,20,25,30,35,40]$ cycles of data experienced.}
  \label{fig:rls_batch}
\end{figure}

\section{Accuracy of Adaptive Geometric Modeling}
\label{sec:adaptive_model_accuracy}

We first seek to demonstrate the speed and precision through which RLS can estimate geometric models of locomotion in real time, provided access to ground truth history of shape and body velocity data. 
We demonstrate the method's performance first on a toy locomotion system, the Purcell swimmer \cite{purcell1977life}, whose local connection can be analytically derived \cite{avron2008geometric,hatton2013geometric}.
This low Reynolds swimmer is a classical example of a principally kinematic system, so its locomotion is explained entirely by our modeling assumptions. 

% We also provide the method's performance on a more complex system using a high-fidelity physical engine. We model an RHex hexapod robot \cite{saranli2001rhex} running in gazebo simulation, which is an example of a principally kinematic system (when operated slowly) that a roboticist is more likely to use in the field today.

Selecting the right data size for the batch model is important in balancing data efficiency and model accuracy, and it is not easily generalizable for arbitrary systems. One advantage of the adaptive model is that it can keep track of the model's accuracy in real time so that there is no need to guess a data size as in the batch model. 
We compared our adaptive models to the batch model mentioned in \ref{datadriven}, in terms of model metrics $\Gamma$ covered in \ref{sec:metrics}. Again, $\Gamma$ captures the knowledge of the model beyond $0$ (the phase-averaged behavior) up to $1$ (full reconstruction of the ground truth). We generated $200$ trials of $40$ cycles of noisy shape trajectories as mentioned in \ref{sec:noise}. We split the data in half, selecting two $40$ cycle trials at a time (100 pairs) and evaluating the model computed on the training set against the held out $40$ cycle test set. In Fig. \ref{fig:rls_batch}, the model accuracies are plotted with respect to the robot's experience throughout the trial. This informs us how the prediction quality grows as the robot's experience grows. The model metrics are shown in boxplots where each box corresponds to $5,25,50,75,95$ percentiles of a model trained with the first $n \in [5,10,15,20,25,30,35,40]$ cycles of data collected and tested on the holdout set. When learning from no history, the adaptive model is capable of obtaining an accuracy of $\Gamma > 0.4$ in 5 cycles, meaning that it can understand forty percent of all perturbations from the limit cycle. This nears the approximate convergent prediction quality of both methods after 40 cycles. The adaptive model metrics are computed continuously, and so are available in between the discretely updated batch model. 

The adaptive model performs relatively well even when there are relatively few samples, whereas the batch model performs best with large datasets. The variation in model quality over the duration of the 40 cycles for the adaptive method speaks to the overfitting produced by the recency bias of the measurements.

Forgetting factors are designed to help weight current (and possibly different) domains more highly, but will also lead the model to potentially overweight data that it has seen more recently. The batch model considers all samples equally, and it is thus less prone to overfitting. Next, we revisit the forgetting factor as a key enabler for the swimmer to regain predictive power on new, unanticipated substrates Fig. \ref{fig:drag_changes}.

\section{Adaptation to Changing Environments}

\begin{figure}[tb]
  \centering
  \includegraphics[width=0.85\columnwidth]{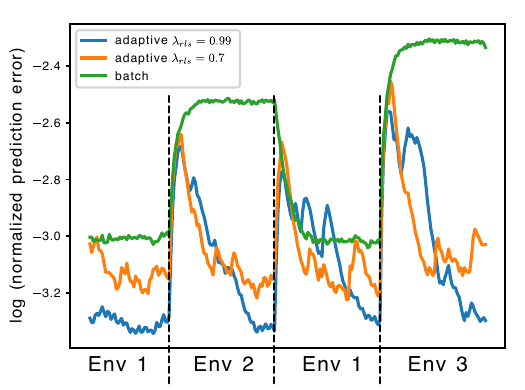}
  \caption{The predictive quality of the adaptive models (blue, $\lambda_{RLS}=0.99$, and orange, ($\lambda_{RLS}=0.7$)) and batch model (green) are compared throughout environmental changes experienced by the Purcell swimmer. Each model is trained with 40 cycles of data collected in `Env 1'. In this test, the system is unknowingly exposed to two other environments, where each environment consists of 40 cycles of data. We changed the drag coefficient ratio experienced on the links for the environment changes, where this ratio is $2.0, 3.0, 4.0$, respectively for environments 1, 2, and 3. The batch model (which is never recomputed during the trial) has constant large predictive errors on new substrates. The adaptive model is refined each time a sample is collected and can adapt to the new environments relatively quickly. We note that the forgetful model has a faster adaptation rate but achieves lower, less stable prediction quality throughout the trial.}
  \label{fig:drag_changes}
\end{figure}

In the derivation of the Purcell swimmer's connection matrix (which defines its mapping from shape velocity to body velocity), the ratio of lateral and longitudinal drag coefficients \cite{cox1970motion} directly affects the nature of locomotion.
When the drag ratio is $1$, the system is unable to exploit the asymmetry of drag to create motion and is unable to displace its center of mass. 
When the drag ratio approaches infinite, each link acts like a non-slip wheel, allowing it to move farther per cycle than when the drag ratio is classically set to $2$.

By altering this drag ratio, we are able to change the physical interactions and test our adaptive method.
We trained the batch and adaptive models with data sampled at the classical drag ratio and then adjusted the ratio over a series of interventions.
We track the speed and precision with which each model adapts in Figure \ref{fig:drag_changes}.
The log prediction error shows that the adaptive model is able to converge to a higher accuracy steady state prediction and does so more quickly than the batch model.
The forgetting factor modulates the rate of adaptation, but in this trial, there appears to be a relationship between adaptation rate and overfitting of the model, which is consistent with our discussion of forgetting factors in \ref{sec:adaptive_model_accuracy}.

\section{In-situ Gait Optimization with Adaptive Geometric Modeling}

\begin{figure}[tb]
  \begin{center}
    \vspace{1mm}
    \includegraphics[width=\columnwidth]{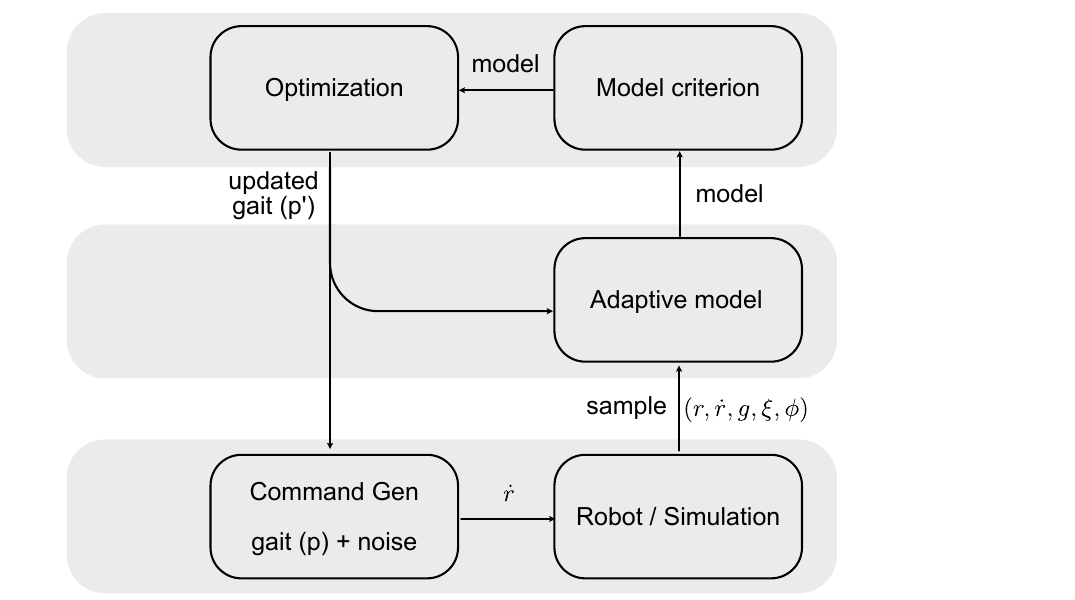}
  \end{center}
  \caption{Above we lay out an optimization architecture for behavior refinement. 
  (Top Level) The optimization will attempt to improve behaviors only when it receives a high enough prediction confidence from the adaptive model.
  (Middle Level) The adaptive model recursively updates from real-world interactions, also producing an estimate of its model confidence in real-time (quantified as $\Gamma$).
  (Bottom Level) A nominal gait, parameterized as p, is given to the \textit{Command Gen} module, which generates stochastic perturbations around the nominal gait.}
  \label{fig:pipeline}
\end{figure}

Here we show that sample-sample refinements of the adaptive model allow for accelerated exploitation of policy gradients for rapid behavior optimization.
A gait optimization in the experiments considered starts with an initial nominal gait, where the system is commanded to move in the neighborhood of the nominal gait using noise signals generated online (see sec. \ref{sec:noise}).
In past work, gait optimization for kinematic systems involved collecting a 30-cycle model, building a batch model (identical to that used as a baseline in Section \ref{sec:adaptive_model_accuracy}), and using that model to step along a policy gradient in the gait parameter space.
30 iterations typically resulted in convergence for the nine-link Purcell swimming totaling 900 total cycles \cite{Bittner2018}.
In this work, we update the adaptive model until we reach an accuracy threshold of $\Gamma=\frac{1}{2}$ (taken over all prior cycles of the current iteration), and then step along the gradient of the policy to select the next set of gait parameters to evaluate.
By this architecture, we can ensure that we use the adaptive models to refine the behavior as soon as they provide an accurate landscape of the local dynamics.
Since each iteration of the optimization only locally changes the dynamics, we expect the RLS filters to smoothly recalibrate the model weights to adapt to the shifted domain of the shape space.
In particular, we managed to obtain faster convergence by updating the zeroth order model (the first element in each RLS weights $\bm{w}[0]$) using previous model predictions on the new gait. 
\begin{equation}
  \label{eq_gait_update}
  % \bm{w}_k^m[0] = [1, \Delta\theta^m, \Delta\dot{\theta}^m, \Delta\theta^m \otimes \Delta\dot{\theta}^m] \cdot \bm{w}_k^m
  \bm{w}_k'[0] = [1, \bm{0}, \bm{0}, \bm{0}] \cdot \bm{w}_k' = [1, \Delta\theta, \Delta\dot{\theta}, \Delta\theta \otimes \Delta\dot{\theta}] \cdot \bm{w}_k 
\end{equation}
Here, the subscript $k$ is the $k^{\mathrm{th}}$ body velocity, $\Delta\theta = \theta' - \theta$ is the difference between the new and old nominal gaits, and $\bm{w}_k'[0]$ is the first element of the model weights (which captures the phase averaged behavior of the new nominal gait).
The rest of the weights are left unchanged because we assume that the new nominal gait shares some variational relationships with the prior gait due to their proximity in the shape space.
This provides a seeded model which we assume will require less refinement for the RLS adaptation.
This zeroth-order term update is equivalent to a Taylor polynomial rebase.
The full optimization pipeline is shown in Fig. \ref{fig:pipeline}.

\begin{figure}
  \centering
  \vspace{3mm}
  \includegraphics[width=0.9\columnwidth]{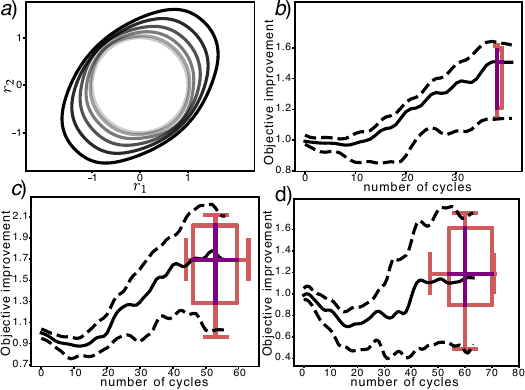}
  \caption{Gait optimization results for Purcell swimmers using the adaptive model optimizer. (A) The three-link Purcell swimmer can refine its gaits from grey to black, optimizing in 40 cycles of experience. 50 trials of the (B) three-link Purcell, (C) five-link Purcell swimmer, and (D) nine-link Purcell swimmer optimization for extremality, with current progress plotted throughout the experience of the trial. Relative performance in displacement per cycle is shown compared to the initial behavior, with [5,50,95] percentiles of performance plotted throughout and [5,25,50,75,95] percentile boxplots provided for the end of the run. The width of the boxplot indicates the number of cycles executed before reaching the final performance value.}
  \label{fig:optimization_shape}
\end{figure}

Figure \ref{fig:optimization_shape} covers an application of this architecture to the Purcell swimmer. 
We repeat the optimization 50 times such that we can observe the range of performance and the repeatability of the process. 
Each swimmer is seeded with an initial gait that is a first-order Fourier series with equal phase lags between each link.
The three-link swimmer is capable of quickly accessing $\Gamma=\frac{1}{2}$ predictive insight and continuously refines its gait over the 40 cycles, improving its seed behavior by almost $50\%$.
The nine-link swimmer, while less reliable than the three-link swimmer, experiences trials in its top quartile that exhibit over $80\%$ improvement of the gait in just 60 cycles.
Our proposed method appears to exhibit a 10-factor improvement in sample efficiency over the work of \cite{Bittner2018}.
Notably, the adaptations are not exclusively successful on the nine-link system, as they were in prior work. 
For more complex systems, careful consideration of appropriate $\Gamma$ and $\lambda_{RLS}$ may improve the reliability of optimization at the expense of sample efficiency.

\section{Discussion and Conclusion}

In this paper, we illustrate the real-time capabilities of our adaptive methods for constructing local models for \textit{in-situ} motion control and behavior refinement.
One immediate advantage of integrating adaptive features is the model's capacity to autonomously adjust in real-time to unanticipated events such as variations in terrain or payload.
The prior batch modeling was unreliable at achieving this real-time capability due to the lengthy requirement to run a batch udpate at every iteration. 
This real-time enhancement brings fast, precise \textit{in-situ} modeling to the field, which has classically been a challenge for platforms with poor first principles models as well as systems that encounter unanticipated scenarios (i.e.,\ not accounted for in training sets).
Unlike reinforcement learning agents and finely calibrated, explicit physics models, our adaptive scheme is designed to refine models quickly and accurately with no assumptions about morphology, weight distribution, or environmental characteristics (so long as the behavior remains principally kinematic).
Without adaptive control, recent work shows that redundancy in the robot body can improve robustness to rugged terrain \cite{chong2023multilegged}. 
A compelling study might involve synthesis of these design principles with the adaptation schemes discussed in this work.

We demonstrated this by showing the ability of the adaptive RLS technique to recapture an accurate physical model under significant substrate variation. 
The adaptation outperformed prior work in terms of speed and data efficiency on the Purcell swimmer as shown in Figures \ref{fig:rls_batch} and \ref{fig:drag_changes}.
%  and simulated Rhex platform. 
These results indicate that the proposed method is an asset for attaining effective motion control in dynamic environments. 
% Based on our results in this work, we can reasonably hypothesize (and seek to test) that the Rhex platform can encounter challenging physical substrates and regain precise motion tracking in a few cycles.
We noted the relationship between forgetting factors, the rate of adaptation, and the consequences of overfitting as observed in our results.
Expectedly, rapid forgetting factors yielded the fastest response to new substrates but the least stable performance. 
Having a dial to tune can be crucial in settings where either (a) speed is critical to success and risk is tolerable (b) speed is not important and safety is paramount.

We presented an optimization architecture where the system executes stochastically perturbed repetitions of the same gait until it gains requisite confidence in its data-driven model of the local dynamics. When this confidence is met, it leads to high-frequency steps along a computed policy gradient, using this data-driven model to select new gaits that locally improve some performance criteria. This led to a significant performance boost for the state of the art, where the Purcell swimmer optimized its gait to an $80\%$ improvement within 60 cycles in its top quartile of performance. It is likely that a large range of the swimmer behavior is attributed to a high forgetting factor $\lambda_{RLS}$ and admissible model prediction threshold $\Gamma$, which could interact to give false confidence, resulting in suboptimal steps along the policy gradient.  Principled selection of these parameters will likely yield different outcomes, such as a less rapid optimizer that converges with near certainty to a similar behavior. 

The gait optimization result presented here demonstrated the method's adaptability to gait changes, involving shifts in the sampling region.  We note that the rebasing of our model coefficients across discrete changes in the behavior space was crucial to accelerating the model adaptation rate.  Notably, we acknowledge that it could also reinforce an overfit model.  However, across many trials, this rebase accelerates the model's capability to explore uncharted regions in the gait space, enabling systematized integration of control, modeling, and optimization processes in real-time with varying physical interactions.  In future work, we seek to integrate these models into model predictive control architectures \cite{williams2017model,carron2019data,hogan2020reactive} to demonstrate refined trajectory tracking that persists through unanticipated scenarios.

\clearpage

\begin{changemargin}{-0cm}{0.05cm}
\IEEEtriggeratref{18}

\bibliographystyle{IEEEtran}
\bibliography{reference}

\begin{thebibliography}{10}
\providecommand{\url}[1]{#1}
\csname url@rmstyle\endcsname
\providecommand{\newblock}{\relax}
\providecommand{\bibinfo}[2]{#2}
\providecommand\BIBentrySTDinterwordspacing{\spaceskip=0pt\relax}
\providecommand\BIBentryALTinterwordstretchfactor{4}
\providecommand\BIBentryALTinterwordspacing{\spaceskip=\fontdimen2\font plus
\BIBentryALTinterwordstretchfactor\fontdimen3\font minus \fontdimen4\font\relax}
\providecommand\BIBforeignlanguage[2]{{%
\expandafter\ifx\csname l@#1\endcsname\relax
\typeout{** WARNING: IEEEtran.bst: No hyphenation pattern has been}%
\typeout{** loaded for the language `#1'. Using the pattern for}%
\typeout{** the default language instead.}%
\else
\language=\csname l@#1\endcsname
\fi
#2}}

\bibitem{grandia2021multi}
R.~Grandia, A.~J. Taylor, A.~D. Ames, and M.~Hutter, ``Multi-layered safety for legged robots via control barrier functions and model predictive control,'' in \emph{International Conference on Robotics and Automation (ICRA)}.\hskip 1em plus 0.5em minus 0.4em\relax IEEE, 2021, pp. 8352--8358.

\bibitem{gong2022zero}
Y.~Gong and J.~W. Grizzle, ``Zero dynamics, pendulum models, and angular momentum in feedback control of bipedal locomotion,'' \emph{Journal of Dynamic Systems, Measurement, and Control}, vol. 144, no.~12, p. 121006, 2022.

\bibitem{weingarten2004automated}
J.~D. Weingarten, G.~A. Lopes, M.~Buehler, R.~E. Groff, and D.~E. Koditschek, ``Automated gait adaptation for legged robots,'' in \emph{International Conference on Robotics and Automation}, vol.~3.\hskip 1em plus 0.5em minus 0.4em\relax IEEE, 2004, pp. 2153--2158.

\bibitem{brambilla2006adaptive}
G.~Brambilla, J.~Buchli, and A.~J. Ijspeert, ``Adaptive four legged locomotion control based on nonlinear dynamical systems,'' in \emph{International Conference on Simulation of Adaptive Behavior}.\hskip 1em plus 0.5em minus 0.4em\relax Springer, 2006, pp. 138--149.

\bibitem{sun2016adaptive}
C.~Sun, W.~He, W.~Ge, and C.~Chang, ``Adaptive neural network control of biped robots,'' \emph{Transactions on Systems, Man, and Cybernetics: systems}, vol.~47, no.~2, pp. 315--326, 2016.

\bibitem{peng2017deeploco}
X.~B. Peng, G.~Berseth, K.~Yin, and M.~Van De~Panne, ``Deeploco: Dynamic locomotion skills using hierarchical deep reinforcement learning,'' \emph{ACM Transactions on Graphics (TOG)}, vol.~36, no.~4, pp. 1--13, 2017.

\bibitem{heess2017emergence}
N.~Heess, D.~Tb, S.~Sriram, J.~Lemmon, J.~Merel, G.~Wayne, Y.~Tassa, T.~Erez, Z.~Wang, S.~Eslami, \emph{et~al.}, ``Emergence of locomotion behaviours in rich environments,'' \emph{arXiv preprint arXiv:1707.02286}, 2017.

\bibitem{tan2018sim}
J.~Tan, T.~Zhang, E.~Coumans, A.~Iscen, Y.~Bai, D.~Hafner, S.~Bohez, and V.~Vanhoucke, ``Sim-to-real: Learning agile locomotion for quadruped robots,'' \emph{arXiv preprint arXiv:1804.10332}, 2018.

\bibitem{tobin2017domain}
J.~Tobin, R.~Fong, A.~Ray, J.~Schneider, W.~Zaremba, and P.~Abbeel, ``Domain randomization for transferring deep neural networks from simulation to the real world,'' in \emph{IEEE/RSJ international conference on intelligent robots and systems (IROS)}.\hskip 1em plus 0.5em minus 0.4em\relax IEEE, 2017, pp. 23--30.

\bibitem{mullins2018adaptive}
G.~E. Mullins, P.~G. Stankiewicz, R.~C. Hawthorne, and S.~K. Gupta, ``Adaptive generation of challenging scenarios for testing and evaluation of autonomous vehicles,'' \emph{Journal of Systems and Software}, vol. 137, pp. 197--215, 2018.

\bibitem{rambabu2019mixture}
R.~Rambabu, P.~Vadakkepat, K.~C. Tan, and M.~Jiang, ``A mixture-of-experts prediction framework for evolutionary dynamic multiobjective optimization,'' \emph{IEEE transactions on cybernetics}, vol.~50, no.~12, pp. 5099--5112, 2019.

\bibitem{tsuda2020modeling}
B.~Tsuda, K.~M. Tye, H.~T. Siegelmann, and T.~J. Sejnowski, ``A modeling framework for adaptive lifelong learning with transfer and savings through gating in the prefrontal cortex,'' \emph{Proceedings of the National Academy of Sciences}, vol. 117, no.~47, pp. 29\,872--29\,882, 2020.

\bibitem{rozen2021design}
S.~Rozen-Levy, W.~Messner, and B.~A. Trimmer, ``The design and development of branch bot: a branch-crawling, caterpillar-inspired, soft robot,'' \emph{The International Journal of Robotics Research}, vol.~40, no.~1, pp. 24--36, 2021.

\bibitem{pantula2022untethered}
A.~Pantula, B.~Datta, Y.~Shi, M.~Wang, J.~Liu, S.~Deng, N.~J. Cowan, T.~D. Nguyen, and D.~H. Gracias, ``Untethered unidirectionally crawling gels driven by asymmetry in contact forces,'' \emph{Science Robotics}, vol.~7, no.~73, p. eadd2903, 2022.

\bibitem{li2020progress}
M.~Li, N.~Xi, Y.~Wang, and L.~Liu, ``Progress in nanorobotics for advancing biomedicine,'' \emph{Transactions on Biomedical Engineering}, vol.~68, no.~1, pp. 130--147, 2020.

\bibitem{park2022computational}
C.~Park, C.~Ozturk, and E.~T. Roche, ``Computational design of a soft robotic myocardium for biomimetic motion and function,'' \emph{Advanced Functional Materials}, vol.~32, no.~40, p. 2206734, 2022.

\bibitem{webster2022biohybrid}
V.~A. Webster-Wood, M.~Guix, N.~W. Xu, B.~Behkam, H.~Sato, D.~Sarkar, S.~Sanchez, M.~Shimizu, and K.~K. Parker, ``Biohybrid robots: Recent progress, challenges, and perspectives,'' \emph{Bioinspiration \& Biomimetics}, 2022.

\bibitem{Bittner2018}
B.~Bittner, R.~L. Hatton, and S.~Revzen, ``Geometrically optimal gaits: a data-driven approach,'' \emph{Nonlinear Dynamics}, vol.~94, no.~3, pp. 1933--1948, 2018.

\bibitem{kelly1995geometric}
S.~D. Kelly and R.~M. Murray, ``Geometric phases and robotic locomotion,'' \emph{Journal of Robotic Systems}, vol.~12, no.~6, pp. 417--431, 1995.

\bibitem{zhao2022walking}
D.~Zhao, B.~Bittner, G.~Clifton, N.~Gravish, and S.~Revzen, ``Walking is like slithering: A unifying, data-driven view of locomotion,'' \emph{Proceedings of the National Academy of Sciences}, vol. 119, no.~37, p. e2113222119, 2022.

\bibitem{chong2023self}
B.~Chong, J.~He, S.~Li, E.~Erickson, K.~Diaz, T.~Wang, D.~Soto, and D.~I. Goldman, ``Self-propulsion via slipping: Frictional swimming in multilegged locomotors,'' \emph{Proceedings of the National Academy of Sciences}, vol. 120, no.~11, p. e2213698120, 2023.

\bibitem{dengenhancing2022}
S.~Deng, R.~L. Hatton, and N.~J. Cowan, ``Enhancing maneuverability via gait design,'' in \emph{Proc IEEE Int Conf Robot Autom}, 2022, pp. 5799--5805.

\bibitem{cendra2001geometric}
H.~Cendra, J.~E. Marsden, and T.~S. Ratiu, ``Geometric mechanics, lagrangian reduction, and nonholonomic systems,'' \emph{Mathematics Unlimited—2001 and Beyond}, pp. 221--273, 2001.

\bibitem{BlochNonholonomic2003}
A.~M. Bloch, \emph{Nonholonomic Mechanics and Control}.\hskip 1em plus 0.5em minus 0.4em\relax Springer New York, NY, 2003.

\bibitem{ostrowskioptimal1998}
J.~P. Ostrowski, J.~P. Desai, and V.~Kumar, ``Optimal gait selection for nonholonomic locomotion systems,'' \emph{The International Journal of Robotics Research}, vol.~19, no.~3, pp. 225--237, 2000.

\bibitem{kelly1996geometry}
S.~D. Kelly and R.~M. Murray, ``The geometry and control of dissipative systems,'' in \emph{Proceedings of 35th IEEE Conference on Decision and Control}, vol.~1.\hskip 1em plus 0.5em minus 0.4em\relax IEEE, 1996, pp. 981--986.

\bibitem{marsden1998symmetries}
J.~E. Marsden and J.~Ostrowski, ``Symmetries in motion: Geometric foundations of motion control,'' \emph{National Academies Press}, 1998.

\bibitem{kvalheim2019gait}
M.~D. Kvalheim, B.~Bittner, and S.~Revzen, ``Gait modeling and optimization for the perturbed stokes regime,'' \emph{Nonlinear Dynamics}, vol.~97, pp. 2249--2270, 2019.

\bibitem{Bittnersuds2022}
B.~Bittner, R.~L. Hatton, and S.~Revzen, ``Data-driven geometric system identification for shape-underactuated dissipative systems,'' \emph{Bioinspiration \& Biomimetics}, vol.~17, no.~2, p. 026004, 2022.

\bibitem{deng2023data}
S.~Deng, J.~Liu, B.~Datta, A.~Pantula, D.~H. Gracias, T.~D. Nguyen, B.~A. Bittner, and N.~J. Cowan, ``A data-driven approach to geometric modeling of systems with low-bandwidth actuator dynamics,'' \emph{arXiv preprint arXiv:2307.01062}, 2023.

\bibitem{hayes1996statistical}
M.~H. Hayes, \emph{Statistical digital signal processing and modeling}.\hskip 1em plus 0.5em minus 0.4em\relax John Wiley \& Sons, 1996.

\bibitem{purcell1977life}
E.~M. Purcell, ``Life at low reynolds number,'' \emph{American journal of physics}, vol.~45, no.~1, pp. 3--11, 1977.

\bibitem{avron2008geometric}
J.~E. Avron and O.~Raz, ``A geometric theory of swimming: Purcell's swimmer and its symmetrized cousin,'' \emph{New Journal of Physics}, vol.~10, no.~6, p. 063016, 2008.

\bibitem{hatton2013geometric}
R.~L. Hatton and H.~Choset, ``Geometric swimming at low and high reynolds numbers,'' \emph{IEEE Transactions on Robotics}, vol.~29, no.~3, pp. 615--624, 2013.

\bibitem{cox1970motion}
R.~G. Cox, ``The motion of long slender bodies in a viscous fluid part 1. general theory,'' \emph{Journal of Fluid Mechanics}, vol.~44, no.~4, p. 791–810, 1970.

\bibitem{chong2023multilegged}
B.~Chong, J.~He, D.~Soto, T.~Wang, D.~Irvine, G.~Blekherman, and D.~I. Goldman, ``Multilegged matter transport: A framework for locomotion on noisy landscapes,'' \emph{Science}, vol. 380, no. 6644, pp. 509--515, 2023.

\bibitem{williams2017model}
G.~Williams, A.~Aldrich, and E.~A. Theodorou, ``Model predictive path integral control: From theory to parallel computation,'' \emph{Journal of Guidance, Control, and Dynamics}, vol.~40, no.~2, pp. 344--357, 2017.

\bibitem{carron2019data}
A.~Carron, E.~Arcari, M.~Wermelinger, L.~Hewing, M.~Hutter, and M.~N. Zeilinger, ``Data-driven model predictive control for trajectory tracking with a robotic arm,'' \emph{Robotics and Automation Letters}, vol.~4, no.~4, pp. 3758--3765, 2019.

\bibitem{hogan2020reactive}
F.~R. Hogan and A.~Rodriguez, ``Reactive planar non-prehensile manipulation with hybrid model predictive control,'' \emph{The International Journal of Robotics Research}, vol.~39, no.~7, pp. 755--773, 2020.

\end{thebibliography}

\end{changemargin}

\end{document}